\documentclass[runningheads,orivec]{llncs}
\usepackage[T1]{fontenc}
\usepackage{hyperref}
\usepackage{graphicx}
\usepackage{color}

\usepackage[caption=false]{subfig}
\usepackage{algorithm2e}
\usepackage{dsfont}
\usepackage{amssymb}
\usepackage{amsmath}
\usepackage{multirow}

\usepackage[normalem]{ulem} %strikethrough

\begin{document}
\title{\textnormal{This is the preprint of the paper that will appear in conference proceedings of Artificial Intelligence in Healthcare (AIiH) 2025}\\[0.5em]GNN's Uncertainty Quantification using Self-Distillation}
\titlerunning{GNN's Uncertainty Quantification using Self-Distillation}

% \author{Anonymous Authors}
% \authorrunning{A. Author et al.}
% \institute{Anonymous Institution}

\author{Hirad Daneshvar\inst{1}*\orcidID{0000-0002-6210-9306} \and
Reza Samavi\inst{1,2}\orcidID{0000-0001-6768-0168}}
\authorrunning{H. Daneshvar \& R. Samavi}
\institute{Toronto Metropolitan University, Toronto, ON, Canada
\email{\{hirad.daneshvar,samavi\}@torontomu.ca}\\ \and
Vector Institute, Toronto, ON, Canada}

\maketitle              % typeset the header of the contribution
\begin{abstract}
%\rs {the abstract needs to be rewritten. after you review the entire paper and eliminate redundancies, I will rewrite the abstract for you} 
Graph Neural Networks (GNNs) have shown remarkable performance in the healthcare domain. However, what remained challenging is quantifying the predictive uncertainty of GNNs, which is an important aspect of trustworthiness in clinical settings. 
%\rs {avoid redundancy: Quantifying uncertainty is especially important as it will provide clinicians with a tool to decide which prediction they can trust.} 
% \rs {While Bayesian methods (e.g., McDropout,...) can be used to quantify uncertainty,......say about their inefficiencies}
While Bayesian and ensemble methods can be used to quantify uncertainty, they are computationally expensive. Additionally, the disagreement metric used by ensemble methods to compute uncertainty cannot capture the diversity of models in an ensemble network. In this paper, we propose a novel method, based on knowledge distillation, to quantify GNNs' uncertainty more efficiently and with higher precision. We apply self-distillation, where the same network serves as both the teacher and student models, thereby avoiding the need to train several networks independently. To ensure the impact of self-distillation, we develop an uncertainty metric 
% \rs {metric of what? diversity metric?} 
that captures the diverse nature of the network by assigning different weights to each GNN classifier. 
% \rs {said nothing about disagreement} \rs{truncated flow} 
We experimentally evaluate the precision, performance, and ability of our approach in distinguishing out-of-distribution data on two graph datasets: MIMIC-IV and Enzymes. The evaluation results demonstrate that the proposed method can effectively capture the predictive uncertainty of the model while having performance similar to that of the MC Dropout and ensemble methods. The code is publicly available at~\url{https://github.com/tailabTMU/UQ_GNN}.
%\footnote{\url{https://github.com/tailabTMU/UQ_GNN}}.

%Graph Neural Networks (GNNs) have shown remarkable performance in the healthcare domain. However, quantifying the predictive uncertainty of GNNs, which is an important aspect of trustworthiness, is challenging. Quantifying uncertainty is especially important as it will provide clinicians with a tool to decide which prediction they can trust. In this research, we propose quantifying the uncertainty of GNNs through self-distillation, an efficient method to train multiple GNN classifiers simultaneously. We train a diverse set of student and teacher models simultaneously using self-distillation. We propose a metric to quantify the predictive uncertainty of the model that can capture the diverse nature of the network by assigning different weights to each GNN classifier. We experimentally evaluated the performance and the capability of our approach in distinguishing out-of-distribution data on two graph datasets: MIMIC-IV and Enzymes. The evaluation results demonstrate that the proposed method can effectively capture the predictive uncertainty of the model while having performance similar to that of the MC Dropout approach and an ensemble of GNN classifiers. The code is publicly available at this URL\footnote{Link will be placed after acceptance.}.

\end{abstract}
\section{Introduction}\label{sec:intro}
% The use of Graph Neural Networks (GNNs) has seen an increase in the medical domain for tasks including diagnosis~\cite{liu2020heterogeneous}, medication recommendation~\cite{shang2019pretraining}, disease prediction, and readmission and mortality prediction~\cite{10.1145/3450439.3451855,Saggu2024}. To effectively use GNNs in healthcare, clinicians need to be able to trust the model predictions~\cite{daneshvar2024sok}. Quantifying the predictive uncertainty of GNNs is crucial for building user's confidence in predictions of the models, especially in sensitive areas like healthcare~\cite{wang2024uncertaintygraphneuralnetworks}. A model's predictive uncertainty score can help clinicians determine which predictions are reliable.
The use of Graph Neural Networks (GNNs) has seen an increase in the medical domain for tasks including diagnosis~\cite{liu2020heterogeneous}, medication recommendation~\cite{shang2019pretraining}, disease prediction, and readmission and mortality prediction~\cite{10.1145/3450439.3451855,Daneshvar2022Heterogeneous}. To effectively use GNNs in healthcare, clinicians need to trust the model predictions~\cite{daneshvar2024sok}. Quantifying the predictive uncertainty of GNNs is crucial for building users' confidence in predictions of the models, especially in sensitive areas like healthcare~\cite{wang2024uncertaintygraphneuralnetworks}.
% A model's predictive uncertainty score can help clinicians determine which predictions are reliable.

% Uncertainty quantification in Neural Networks has recently seen significant interest, with Bayesian methods being a prominent approach.  While Bayesian methods offer precise uncertainty estimates by learning a distribution over model parameters, they suffer from two key drawbacks: high computational cost and the need for substantial model modifications. As a more practical alternative, training an ensemble of GNNs has been explored for uncertainty quantification and the disagreement between individual model predictions, in which the ensemble's outcome is used to quantify predictive uncertainty~\cite{NIPS2017_9ef2ed4b}. 
Bayesian methods provide precise uncertainty quantification in neural networks but are computationally expensive.
%and require significant changes in the model. 
Alternatively, ensemble methods quantify uncertainty by measuring disagreement among multiple single network predictions and the ensemble outcome as the reference~\cite{NIPS2017_9ef2ed4b}. To ensure diversity in the ensemble, knowledge distillation is employed, where a teacher model is first trained and then its predictions are used to guide the training of smaller student models~\cite{10.1145/3318464.3389706}. 
Despite the improved scalability, the ensemble approach remains computationally expensive, particularly with deep GNNs, as several deep individual networks need to be trained. Furthermore, the disagreement metric can be imprecise when the ensemble contains diverse models. The following example further illustrates the limitations of the disagreement metric.

Consider a clinical institute that employs three GNN classifiers, C1, C2 (deeper than C1), and C3 (deeper than C2), to diagnose Common Cold, COVID-19, and Influenza. Predictions from C3 serve as the reference. C1 and C2 are used to assess the uncertainty of predictions. If all models agree on COVID-19 (Patient~1), uncertainty is low; when C1 and C2 predict COVID-19 but C3 predicts Influenza (Patient~2), uncertainty should be higher. However, the disagreement metric assigns the same uncertainty (0.070) to both cases (Table~\ref{tab:motiv_exam_1}), demonstrating the lack of expressivity of the current disagreement metric. 

\begin{table}[h]
\centering
\vspace{-.5cm}
    \caption{Disagreement and classification by three GNNs (with increasing depth from C1 to C3) for two patients, highlighting the disagreement metric's limited informativeness when shallower GNNs (C1 and C2) deviate from the Reference GNN (C3).}\label{tab:motiv_exam_1}
    % Three GNN classifiers (GNN~1, GNN~2, Ref. GNN, ordered by increasing depth) show disagreement and classification results for two patients. GNN~1 and GNN~2, being shallower, aim to mimic Ref. GNN. The disagreement metric is less informative when GNN~1 and GNN~2 disagree with Ref. GNN's predictions.}\label{tab:motiv_exam_1}
            \begin{tabular}{|c|c|c|c|c|c|}
            \hline
                 \textbf{Patients}&\textbf{Model}& \textbf{Common Cold}& \textbf{Covid-19}& \textbf{Influenza} &\textbf{Disagreement}\\
                \hline
              \multirow{3}{*}{\textbf{Patient~1}}&C1& 0.3& \textit{0.38}& 0.32 & \multirow{3}{*}{0.070}\\
              &C2& 0.3& \textit{0.4}& 0.3 &\\
              &C3& 0.2& \textit{0.5}& 0.3 &\\
             \hline
              \multirow{3}{*}{\textbf{Patient~2}}&C1
            & 0.301& \textit{0.414}& 0.286 & \multirow{3}{*}{0.070}\\
              &C2 & 0.301& \textit{0.442}& 0.257 &\\
              &C3& 0.3& 0.32& \textit{0.38} &\\   
              \hline
            \end{tabular}
            \vspace{-.2cm}
\end{table}

% \rs {This sentence is problematic - do not use This is. also briefly define what you mean by Harder example. I tried to fix it but needs a bit of work from you, then I'll fix it.}

\noindent In this paper, we propose a self-distillation-based approach for efficient and precise uncertainty quantification of GNNs. Self-distillation represents a particular kind of knowledge distillation where the teacher and student models are part of a single model~\cite{9381661}. 
% \rs {here you have to informally define in one sentence what is self-distillaiton. you did it in abstract but missed it here!}
Efficiency is achieved through deepest classifier teacher distillation in a multi-classifier GNN. During training, the shallower classifiers learn from the deepest classifier using a combined cross-entropy and Kullback–Leibler (KL) divergence loss. The precision is improved by a weighted disagreement metric that emphasizes the diversity of the classifiers. The  metric assigns higher weights to deeper classifiers when their predictions diverge from those of the teacher. Since deeper classifiers utilize richer representations, a discrepancy between the teacher and deeper classifiers for classifying a data point indicates that the classifier is having difficulty aligning itself with the teacher (i.e., the deepest) classifier, causing the data point to be considered as a hard example~\cite{9381661}.

The key contributions of this paper are as follows. First, we develop a self-distillation-based method for uncertainty quantification that is more efficient than training separate individual GNNs in an ensemble. Second, we propose a refined uncertainty metric that demonstrates greater precision by incorporating diversity among individual networks of an ensemble. Finally, we experimentally demonstrate the efficiency of our approach in quantifying GNN uncertainty.
% Finally, we experimentally demonstrate how our approach can be applied to quantify GNN uncertainty efficiently.
\section{Related Work}\label{sec:related}

MC Dropout~\cite{Kwon2022} and last-layer dropout~\cite{2022_Aouichaoui} have been applied to GNNs for Bayesian inference, learning parameter distributions and quantifying uncertainty through the variance of multiple predictions. A Bayesian semi-supervised Graph Convolution Neural network~\cite{C9SC00616H} uses predictive variance for uncertainty, particularly in low-data scenarios. Bayesian methods are computationally expensive and require multiple predictions for uncertainty estimation. The proposed approach aims to achieve efficient uncertainty quantification in any GNN classifier.

% MC Dropout for Bayesian inference has also been applied to GNNs~\cite{Kwon2022}. In this approach, the model learns a probability distribution over the parameters of the model. To quantify uncertainty, the predictions are sampled multiple times, and the predictive variance of the sampled predictions is computed as a measure of uncertainty. In a similar approach, the last layer dropout for Bayesian inference was applied to GNNs~\cite{2022_Aouichaoui}. In this approach, the distribution of the predictions is calculated using MC simulation. A Bayesian semi-supervised Graph Convolution Neural network has also been proposed~\cite{C9SC00616H}. The authors utilized semi-supervised learning to ensure that the uncertainty quantification is accurate in scenarios with low data limit. 
%The approach will help with active learning with limited initial data.
% Similar to the previous approaches, the proposed approach uses the variance of the predictions to quantify uncertainty. The Bayesian approach is computationally expensive. Additionally, there is a need to make changes to the model, and several samples of the prediction should be used to compute uncertainty. Our proposed approach can be utilized for any GNN classifier with minimal changes to the training process.

Conformal prediction is a statistical framework that produces prediction intervals for a pretrained model with a guarantee on the prediction’s reliability~\cite{vovk2005algorithmic}. Although integrating conformal interval predictions into distributional methods of uncertainty quantification is challenging, probabilistic methods have been developed to provide certified boundaries of computed uncertainty~\cite{karimi2023quantifying,karimi2024evidential}. Conformalized GNN (CGNN) extends conformal prediction to GNNs for uncertainty quantification~\cite{NEURIPS2023_54a1495b}, by using the output of a GNN to train a second GNN for conformal corrections and finally applying standard conformal prediction to generate prediction sets. Another approach applies conformal prediction directly to pre-trained GNNs using a graph-aware conformity score based on neighbourhood diffusion~\cite{pmlr-v202-h-zargarbashi23a}. These methods primarily focus on node classification, requiring modifications for graph classification. Our proposed approach can be applied to any classification task using GNNs, including node and graph classification.

% Conformalized GNN was proposed to extend conformal prediction to GNNs and quantify uncertainty~\cite{NEURIPS2023_54a1495b}. It first starts by training a GNN. The output of the trained GNN for each node is used as the input to train a second GNN to make conformal corrections and update predictions. As the final step, standard conformal prediction is used to create prediction sets. Using conformal prediction on any pre-trained GNN has also been proposed by~\cite{pmlr-v202-h-zargarbashi23a}. The proposed approach uses a conformity score for each node. The approach uses the graph structure, which updates the conformal score for each node based on neighbourhood diffusion. The proposed conformal prediction-based approaches are mainly developed for node classification tasks. Therefore, generalizing them to graph classification requires changes to the proposed methods. Our proposed approach can be applied to both node and graph classification tasks.

Ensembles can be utilized to quantify model uncertainty by aggregating predictions from multiple models. \cite{NIPS2017_9ef2ed4b}~proposed a general deep ensemble framework to quantify model uncertainty by using the KL divergence of the individual network predictions in the ensemble, and the ensemble outcome. \cite{2022_Aouichaoui}~trained independent GNNs and used the variance of their predictions as an uncertainty measure. The previous approaches mainly focused on training identical networks in an ensemble. Teacher-student knowledge distillation has been utilized to add diversity to the ensemble~\cite{10.1145/3318464.3389706}. \cite{10889349} utilized a GNN with a multi-head classification layer, in which the classification heads utilize the same GNN backbone, and used the variance of prediction to estimate uncertainty. However, training an ensemble is computationally expensive, and existing disagreement metrics may not be suitable for diverse ensembles. Our approach trains classifiers simultaneously with an uncertainty metric that captures the diversity of the models.

\section{Self-Distillation for Uncertainty Estimation}\label{sec:method}
In this section, we will first provide the necessary setup for knowledge distillation and self-distillation. We then present the training phase of the method and the proposed uncertainty quantification metric. We utilize GNNs for the graph classification task throughout this work, even though the approach is generalizable to any GNN's classification task, including graph and node classification.

\subsection{Problem Setup}\label{subsec:method.overview}
\paragraph{Knowledge Distillation} is an effective technique for training a smaller, simpler student model by learning from a larger, more complex teacher model, achieving performance comparable to that of the teacher model while having reduced computational demands. 
% while significantly reducing computational demands. 
The approach facilitates the deployment of neural networks on resource-constrained devices~\cite{hinton2015distillingknowledgeneuralnetwork,Gou2021}. 
%The goal of knowledge distillation is to achieve performance comparable to the teacher model while significantly reducing computational demands.
Traditional knowledge distillation starts by training a teacher model. Subsequently, the teacher's soft output probabilities (soft labels) are used to train one or more separate, typically shallower, student models that aim to replicate the teacher's behavior~\cite{pmlr-v97-phuong19a}. This process effectively transfers knowledge from the teacher to the student. However, the training process can be resource-intensive, requiring the separate training of both the teacher and potentially multiple student models.

\paragraph{Self-Distillation} is a specific type of knowledge distillation~\cite{Gou2021}  that uses the same network as both the teacher and the student and offers a solution for the challenges of knowledge distillation in over-parameterized GNNs~\cite{chen2021selfdistillinggraphneuralnetwork}. 
%As a specific type of knowledge distillation~\cite{Gou2021}, it uniquely uses the same network as both the teacher and the student. 
In contrast to traditional knowledge distillation, self-distillation involves transferring knowledge from the model's deeper layers to its shallower ones~\cite{9381661,Gou2021}. In self-distillation, a classifier is placed after each hidden layer, where the classifier connected to the deepest layer acts as a teacher. For example, a GNN with three hidden layers would have three classifiers, each using the output of its previous layers for classification. Consequently, deeper classifiers utilize more comprehensive representations of the input, having access to features extracted by deeper layers. When hidden layers have different output dimensions, a harmonization layer is applied before all classifiers, except the deepest one. This layer, often a fully connected layer, adjusts the output shape of its preceding hidden layer to match the final hidden layer, ensuring uniform input shapes for all classifiers.

\subsection{Training with Self-Distillation}\label{subsec:method.training}
\begin{figure}[t]
\centering
\includegraphics[width=.67\textwidth]{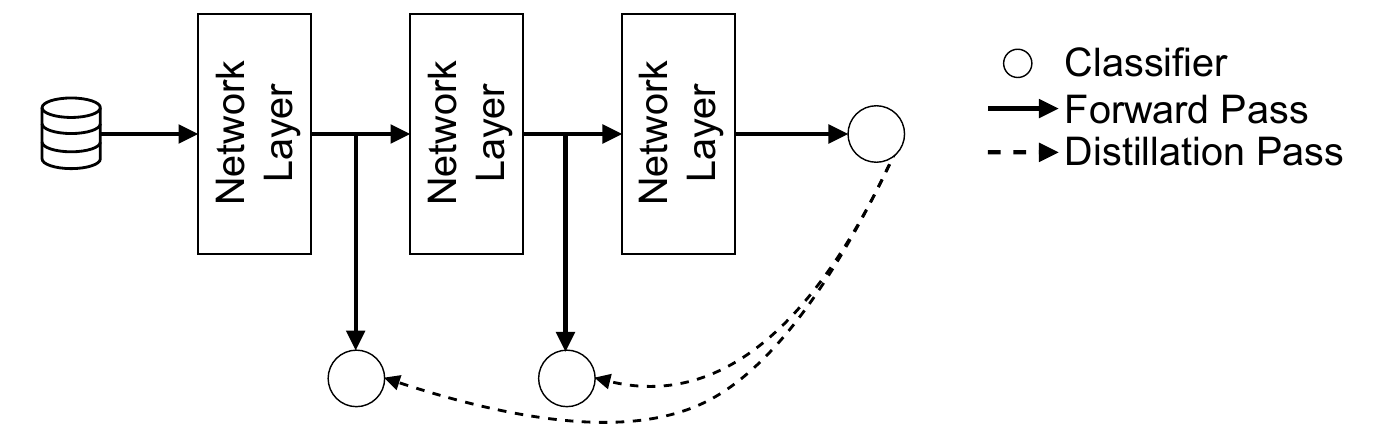} 
\caption{A network structure with 3 layers and 3 classifiers used in self-distillation.}
\label{fig:self_distil}
\end{figure}

% In this work, we first train a GNN model with multiple classifiers using the self-distillation approach. 
Figure~\ref{fig:self_distil} shows a sample neural network with three hidden layers and three classifiers developed for self-distillation. 
% Each classifier uses the output provided by its previous hidden layers for classification.
Let $f:\mathcal{G} \rightarrow \mathcal{Y}$ be a GNN classifier trained on $D=\Big\{(G_i,y_i) | G_i \in \mathcal{G} , y_i \in \mathcal{Y}\Big\}_{i}^{N}$. The network $f$ consists of $m$ hidden layers, i.e., hidden graph operators, each followed by a classifier $c_{l}$, such as a Multilayer Perceptron (MLP), where $1 \leq l \leq m$. Each hidden layer is a feature extractor $h_{l}(G)$ that uses the node features and the adjacency matrix of a graph $G$ to extract graph features. $c_{l}\Big(h_{l}(x_{i})\Big)$ represents the predicted output of the classifier using the extracted features at layer $l$.

During the training process, each classifier learns to mimic the teacher model, i.e., the deepest classifier in Figure~\ref{fig:self_distil}. To this end, a total distillation loss function is used. The total distillation loss is based on the work by~\cite{9381661} and has two components: \textit{distillation loss} and \textit{feature penalty}. The distillation loss minimizes the compound cross-entropy loss and KL divergence loss of every classifier, and the feature penalty minimizes a penalty for features extracted by shallower models. The training objective is to minimize the total distillation loss.

\begin{definition}\label{def:dist_loss}
    For each graph $G_{i}$, the distillation loss $L_{\mathrm{dis}}^i$ is the mean of layer-wise weighted sums of two components: the cross-entropy of the classifier output at layer $l$, $c_{l}(G_{i})$, with the true label, and the KL-divergence between the student's soft labels at layer $l$, $q_{l}^{i}$, and the teacher's soft labels, $q_{t}^{i}$. The loss is computed as

    % For every graph $G_{i}$, the distillation loss $L_{\mathrm{dis}}^i$ is computed by the mean of the layer-wise weighted sum of cross-entropy of $c_{l}(G_{i})$ with respect to the true label and the KL-divergence of soft labels provided by the model at layer $l$, $q_{l}^{i}$, with respect to the soft labels of the teacher model, i.e., $q_{t}^{i}$. The loss is computed as

    \begin{equation}\label{eq:dist_loss}
        L_{\mathrm{dis}}^i = \frac{1}{m} \sum_{l=1}^m \Big( (1-\alpha_l) L_{\mathrm{CE}_{l}}^{i} + \alpha_l L_{\mathrm{KL}_{l}}^{i} \Big),
    \end{equation}
    where $\alpha _{l} \in [0, 1]$ is the imitation parameter for each classifier. The $L_{\mathrm{CE}_{l}}$ and $L_{\mathrm{KL}_{l}}$ are the cross-entropy and KL divergence loss for layer $l$, respectively. The imitation parameter of the teacher model is set to zero. As a result, the teacher model will be trained only using the cross-entropy loss.
\end{definition}

\noindent The cross-entropy loss of the model is formulated as

\begin{equation}\label{eq:cross_entropy_lss}
    \begin{aligned}
    L_{\mathrm{CE}_{l}}^i = -\sum_{c=1}^{k}y_{i}^{c}\log \Big(q_{l}^{i} \Big),
    \end{aligned}
\end{equation}
and the KL divergence loss is formulated as
\begin{equation}\label{eq:kl_div_loss}
L_{\mathrm{KL}_{l}}^i = \sum_{c=1}^{k} q_{l}^{i} \log \Bigg( \frac{q_{l}^{i}}{q_{t}^{i}} \Bigg).
\end{equation}
% \rs {describe the terms for (2) and (3)}
\noindent Similar to~\cite{9381661}, we add a penalty for features extracted by shallower models.

\begin{definition}\label{def:pen:loss}
For each graph $G_{i}$, the penalty $L_{\mathrm{pen}}^i$ is the weighted average of the squared $\ell_2$-norm of the features extracted in each layer, $h_{l}$, and the features extracted by the teacher model, $h_{t}$. The penalty is computed as
    \begin{equation}\label{eq:pen:loss}
    L_{\mathrm{pen}}^i = \frac{1}{m} \sum _{l=1}^m \lambda _{l} L_2 \Big(h_{l}^{i}, h_{t}^{i} \Big),
    \end{equation}
where $\lambda _{l} > 0$ is the trade-off parameter for each classifier and $L_2$ is the squared $\ell_2$-norm loss. The $\lambda_{l}$ parameter for the teacher model is set to zero.
\end{definition}

% The total distillation loss for all $n$ graphs in the training dataset is the combination of both the distillation loss (Definition~\ref{def:dist_loss}), and the penalty component (Definition~\ref{def:pen:loss}), which is

\noindent The total distillation loss for all $n$ graphs in the training dataset is the mean of the combination of both the distillation loss and penalty component, which is

\begin{equation}\label{eq:final_dist_loss}
L = \frac{1}{n} \sum _{i=1}^n L_{\mathrm{dist}}^{i} + L_{\mathrm{pen}}^{i}.
\end{equation}

\subsection{Uncertainty Quantification}\label{subsec:method.uncert-quant}
To quantify uncertainty, we utilize the difference between predictions of each classifier $c_{l}$, where $1 \leq l \leq m-1$, and the teacher classifier $c_{t}$, where $t = m$. Our intuition of uncertainty quantification is rooted in the disagreement metric used to quantify uncertainty in a deep ensemble~\cite{NIPS2017_9ef2ed4b}. This notion can be formally defined as a disagreement between the probability predictions of each individual model and 
% the probability predictions 
that of the ensemble. In other words, in a deep ensemble, if the prediction of each model aligns with the prediction of the ensemble, there would be less prediction uncertainty for the given data.

\begin{definition}\label{def:disagreement}
% The disagreement metric for a graph $G$ in an ensemble of $m$ GNNs can be quantified by summing the KL divergence between the prediction probabilities of each model and the prediction probabilities of the ensemble outcome, which is,
The disagreement metric for a graph $G$ in an ensemble of $m$ GNNs is the sum of KL divergences between each model's prediction and the ensemble's outcome, which is,

\begin{equation}\label{eq:disagreement}
    disagreement = \sum_{l=1}^{m}KL\Big(P_{l}||P_{outcome}\Big),
\end{equation}
where $P_{outcome}$ is the prediction probabilities of the ensemble model.
\end{definition}

\noindent The KL-divergence measures the difference between two probability distributions. When the distribution of both probabilities is the same, the KL-divergence is zero, and the value gets unboundedly larger as the two probability distributions diverge. KL-divergence can be computed as

\begin{equation}\label{eq:kl_div}
\begin{aligned}
KL\Big(P_{l}(y|G)|| P_{outcome}(y|G)\Big) = \sum_{c=1}^{k} P_{l}(y_{c}|G) \log \Bigg( \frac{P_{l}(y_{c}|G)}{P_{outcome}(y_{c}|G)} \Bigg),
\end{aligned}
\end{equation}
where $k$ is the total number of labels.

In self-distillation, every layer is followed by a shallow classifier (e.g., a two-layer MLP), where the preceding layers and the classifier are considered as a student model, except for the deepest layer, which acts as the teacher. 
% Classifiers that utilize features derived from deeper network layers, i.e., deeper classifiers, 
Deeper classifiers utilize richer representations. 
% and exhibit higher accuracy compared to shallow classifiers, which typically demonstrate slightly lower accuracy. 
Therefore, if the predictions of a shallow classifier deviate from those of deeper classifiers, the instance can be considered as a hard example~\cite{9381661}. This distinction arises because deeper classifiers exploit more informative feature representations. 
% If, for a data point, the predictions of the shallower classifiers align with the predictions of the teacher model, the given data point can be considered an easy example. Thus, the model should exhibit less uncertainty.
% As shown in Table~\ref{tab:motiv_exam_1}, the disagreement metric introduced in Equation~\ref{eq:disagreement} treats all models uniformly, especially when there is a disagreement in the predicted class. In other words, the disagreement metric cannot capture the diversity of network depth in the ensemble in scenarios where there is a disagreement between the predicted class of the model and the predicted class of the outcome (reference). Our intuition led us to consider a weight function to break the uniformity of the importance of the models when there are disagreements in the predicted class.
The disagreement metric (Equation~\ref{eq:disagreement}), as shown in Table~\ref{tab:motiv_exam_1}, treats all models equally, especially when predicted classes differ from the outcome of the deepest network, i.e. the reference/teacher model. To capture the impact of network depth diversity in such disagreements, we propose a weight function to differentiate varying depths of the models. %model importance.

\paragraph{Weight Function.} The weight function puts more emphasis on deeper student classifiers when their prediction disagrees with the outcome, thus increasing the impact of their disagreement on the uncertainty value. This function, $W(d)$, is monotonically increasing with depth ($W(1) < W(2) < \dots < W(L)$) and bounded between 1 and 2. If a student's prediction matches the outcome, its weight is 1; otherwise, the weight increases with its depth.
% The weight function captures the diversity in depth of the student classifiers by assigning higher weights to deeper classifiers in case the predicted class of the outcome differs from that of the student classifier. Thus, a disagreement between the outcome and the deeper classifiers has a greater impact on the uncertainty value than a disagreement between the outcome and shallower classifiers. The only constraint for the weight function is to be monotonically increasing to be able to successfully assign higher weights to deeper classifiers, $W(1) < W(2) < W(3) < ... < W(L)$. In the proposed approach, the weights are bounded $1 \leq w \leq 2$. If the predicted class of the student and the outcome are the same, the weight is computed as 1. Otherwise, the weight is computed based on the model's depth.
We use the teacher's prediction as the outcome since all shallower classifiers mimic the teacher model.

The uncertainty of the model for a graph $G$ can be formally quantified as

\begin{equation}\label{eq:weighted_kl_div}
    disagreement_{w} = \sum_{l=1}^{m} W(l) \times KL(P_{l}||P_{teacher}),
\end{equation}
where $W$ is a weight function and $P_{teacher}$ is the predicted probabilities of the teacher model. 
% The weight function needs to utilize the distance from the deepest layer to compute the weight for each classifier. The closer a classifier is to the deepest classifier, i.e., the deeper the feature extractor is, the more weight the weight function should return. 
The weight function assigns higher weights to classifiers based on their depth relative to the deepest classifier. Specifically, classifiers with feature extractors of greater depth receive higher weights. The weight of a classifier at layer $l$ can be computed by

\begin{equation}\label{eq:linear_weight}
    W_{lin}(l) = 1 + \frac{L - D(l)}{L}\mathds{1}_{\{y_{l} \neq y_{teacher}\}},
\end{equation}
where $L$ is the total number of layers in the network, $D(i)$ is the distance of the classifier $i$ from the deepest layer and $\mathds{1}$ is the indicator function, which returns one if the given condition (e.g., if $y_{l}$ differs from $y_{teacher}$) is met; otherwise, it returns zero. $y_l$  and $y_{teacher}$ are the predicted class by the $l$th classifier and the predicted class by the teacher classifier, respectively.

% One limitation of our proposed metric is that we use a linear weight function. 
Alternatively, the weight function can be nonlinear. However, the impact of using a nonlinear function needs further investigation. To show the potential of utilizing a nonlinear weight function, we introduce a nonlinear alternative $1 \leq W_{nonlinear}(l) \leq 2$ that assigns higher weights compared to $W_{linear}(l)$. The nonlinear weight function is

\begin{equation}\label{eq:nonlinear_weight}
    W_{nonlin}(l) = (-(exp(D(l)-L))+2)^{\mathds{1}_{\{y_{l} \neq y_{teacher}\}}},
\end{equation}
where the $exp$ component exponentially amplifies the disagreement effect in deeper layers. Mathematically, both $W_{lin}(l)$ and $W_{nonlin}(l)$ weight functions are sound as they are monotonically increasing. The choice between linear or non-linear weight functions can be considered as a domain-specific parameter to make the uncertainty quantification more or less conservative. Nevertheless, understanding the impact requires further empirical investigation. 
% because they increase the weight assigned as the classifier depth increases.

Weighted disagreement (Equation~\ref{eq:weighted_kl_div}) uses unbounded KL Divergence, which increases significantly with larger differences between $P_{outcome}$ and $P_{l}$, especially near zero probabilities in $P_{outcome}$. To address this, we use bounded Jensen-Shannon Divergence ($0 \leq JSD \leq \log_{b}(2)$), which is based on KL Divergence. Thus, our proposed uncertainty metric can be computed as

% A limitation of the weighted disagreement (Equation~\ref{eq:weighted_kl_div}) is the use of KL Divergence. KL Divergence is unbounded. As the difference between $P_{outcome}$ and $P_{l}$ grows, KL-divergence increases. This is especially true when there are probabilities closer to zero in $P_{outcome}$. To address this limitation, we use the Jensen-Shannon Divergence (JSD). Jensen-Shannon divergence is based on KL Divergence but is bounded $0 \leq JSD \leq \log_{b}(2)$. Our proposed uncertainty metric can be computed as

\begin{equation}\label{eq:uncertainty}
    UC = \sum_{l=1}^{m} W(l) \times JSD(P_{l}||P_{teacher}),
\end{equation}
where the JSD can be computed by

\begin{equation}\label{eq:js_div}
    \begin{aligned}
    JSD\Big(P_{l}||P_{outcome}\Big) = \frac{1}{2} \Bigg(KL(P_{l}||M)+KL(P_{outcome}||M) \Bigg),
    \end{aligned}
\end{equation}
where $M$ is a mixture distribution of $P_{l}$ and $P_{outcome}$ and can be computed as $M=\frac{1}{2}(P_{l}+P_{outcome})$. The mixture distribution helps with averaging and smoothing out the values, which causes the JSD to be bounded.

Although the JSD is bounded, in our approach, the upper bound depends on the number of layers in the network, causing problems in interpreting and comparing the results for networks with different numbers of layers. Therefore, we propose to normalize the $UC$ results by dividing them by the upper bound of the $UC$ metric. The upper bound of $UC$ for a network with $m$ layers (including the final teacher layer) can be computed as

\begin{equation}\label{eq:uc_max}
    UC_{max} = \sum_{l=1}^{m-1} W(l) \times \log_{b}(2),
\end{equation}
where $b$ is the logarithm base used to compute JSD. In our work, we are using the natural logarithm, i.e., $b=e$. $W(l)$ computes the weight at layer $l$. We can finally compute our normalized metric as

\begin{equation}\label{eq:uc_norm}
    UC_{norm} = \frac{UC}{UC_{max}}.
\end{equation}

Algorithm~\ref{alg:uncert_quant_gnn} outlines the steps to compute model uncertainty with our proposed disagreement metric. The algorithm relies on predictions from both the teacher and student models, as shown in line 2. If the student's prediction differs from the outcome of the teacher/reference model, the weight function computes the classifier's weight based on its depth (lines 3 to 12).

% \IncMargin{1em}
\begin{algorithm}[t]
\caption{Quantifying the uncertainty of the GNN prediction}
\label{alg:uncert_quant_gnn}
\DontPrintSemicolon
\LinesNumbered
\SetAlgoLined
\KwIn{Graph input ${x}$, trained GNN $M$, and total number of layers $m$}
\KwOut{Uncertainty value $u$}
$u \leftarrow 0$\;
% \textcolor{blue}{\tcp{Step 1: Get predictions.}}
($preds_{t}, preds_{s}) \leftarrow M(x)$ \textcolor{blue}{\tcp*{Step 1: Get predictions}}
% \textcolor{blue}{\tcp{Step 2: Compute weights.}}
$w \leftarrow [m]$ \textcolor{blue}{\tcp*{Step 2: Compute weights (array of size $m$).}}
$outcome\_class = argmax(preds_{t})$\;
\For{$l\leftarrow 1$ \KwTo $(m-1)$}{
    $student\_class = argmax(preds_{s}[l])$\;
    \uIf{$student\_class \, != \, outcome\_class$}{
        $w[l] \leftarrow 1 + weight\_fn(l, m)$\;
    }
    \Else{
        $w[l] \leftarrow 1 $\;
    }
}
% \textcolor{blue}{\tcp{Step 3: Compute disagreements.}}
$d \leftarrow [m]$ \textcolor{blue}{\tcp*{Step 3: Compute disagreements (array of size $m$).}}
\For{$l\leftarrow 1$ \KwTo $(m-1)$}{
    $d[l] \leftarrow disagreement(preds_{s}[l], preds_{t})$\;
}
% \textcolor{blue}{\tcp{Step 4: Compute uncertainty.}}
$u \leftarrow sum(w \times d)$\ \textcolor{blue}{\tcp*{Step 4: Compute uncertainty.}}
\Return $u$\;
\end{algorithm}
% \IncMargin{1em}

\section{Experimental Evaluations}\label{sec:eval}
% The goal of the proposed approach is to quantify the uncertainty of GNN classifiers precisely and efficiently. We verify its precision by comparing the uncertainty values of the proposed uncertainty metric and the disagreement metric for scenarios in which the predicted class of the outcome and a shallower classifier do not align. We evaluate the efficiency of our approach by comparing the time it takes to train the model and the inference time on the test dataset. We compared the results of a model trained using our approach to a single GNN classifier, the MC Dropout approach, and an ensemble of independent models. Additionally, we compare the three approaches on out-of-distribution (OOD) data to verify that, similar to the deep ensemble and MC Dropout approaches, our approach can capture OOD data. Finally, to measure the possible overhead of our approach to the classification performance, we measure the classification performance of a model trained using our approach, the ensemble, and MC Dropout approaches. This is to verify that the proposed approach does not greatly impact the classification performance of the model.

The proposed approach aims for precise and efficient GNN uncertainty quantification. We assess precision by comparing its uncertainty values with the disagreement metric in cases where there is a mismatch between the predictions of the reference and the shallow classifier. Efficiency is evaluated by comparing training and inference times against a single GNN, MC Dropout, and an independent ensemble. We also test OOD detection capabilities and measure potential classification performance overhead compared to the ensemble and MC Dropout.

\subsection{Experimental Setup}\label{subsec:exp_setup}
% \textbf{Dataset:} We utilized the MIMIC-IV, a publicly available clinical dataset~\cite{Johnson2024-vw,Johnson2023-lj}.
\paragraph{Dataset:} We used the publicly available MIMIC-IV dataset~\cite{Johnson2024-vw}, containing medical data of over 200,000 patients labeled as "Not Readmitted," "Admitted to ICU," or "Admitted for Other Reasons" within 30 days of a hospital visit. For each patient, we constructed a heterogeneous directed graph with \textbf{visit nodes} (admission type, location) and \textbf{service nodes} (service code), following~\cite{Daneshvar2022Heterogeneous}. Edges connect temporally ordered visits and visits to utilized services. 
% A sample patient is shown in Figure~\ref{fig:sample_mimic_data}. 
We used the "Admitted to ICU" as OOD data. For training, we balanced the "Not Admitted" (213,547 patients) and "Admitted for Other Reasons" (7,359 patients) groups by undersampling the former to 7,359 patients. To evaluate generalizability, we also utilized the Enzymes dataset~\cite{Morris+2020}, which comprises 600 graphs, each with an average of 32.6 nodes, each node having 18 features. The dataset contains six classes, where five classes are used for training and the last for OOD detection.

\paragraph{Models:} For the Enzymes dataset, we utilized a graph classifier with four GraphSAGE~\cite{NIPS2017_hamilton} layers in all three approaches. The models follow the GNN structure provided by~\cite{Dwivedi2023}. The ensemble approach trains four identical networks with different random initializations. The MC Dropout approach utilizes a dropout layer before the fully connected layer, and the Self-Distillation approach adds a classifier after each GNN layer. For the MIMIC-IV dataset, we followed the same strategy as the Enzymes dataset, with the only difference being that the networks have three GraphConv~\cite{morris2021weisfeiler} layers and there are three models in the ensemble. We followed the same structure and pooling layer suggested by~\cite{Daneshvar2022Heterogeneous}.
% The details of the models can be found in Appendix~\ref{apd:models}.

\paragraph{Training and Hardware Specifications:} All models were implemented using Python version 3.9 and PyTorch version 1.13.1. The models have been trained on a GPU (NVIDIA GeForce RTX 3050) with CUDA version 12.5. To optimize the networks, we used Adam Optimizer. 
% We only used classes 1 to 5 of the Enzymes dataset and classes 0 and 1 of the MIMIC-IV dataset for training and testing our method. The sixth class of the Enzymes dataset, as well as the third class of the MIMIC-IV dataset, have been used as the OOD data. 
We conducted a stratified 5-fold cross-validation. In each iteration, 80\% of the training data was used for training, and the rest for validation. To aid the training process and prevent overfitting, batch normalization and dropout techniques were used for all models.

% To compare the methods, each model in the ensemble and the model used for the MC Dropout approach were trained separately with the Cross-Entropy Loss function. Then, we trained the model prepared for the self-distillation approach using the loss function provided by Equation~\ref{eq:final_dist_loss}. To maintain consistency and comparability, the same folds and splits of data were used to train the models. For the self-distillation approach, we used 0.6 for $\alpha$ and 0.04 for $\lambda$ as suggested by~\cite{9381661} for all layers except for the reference layer. Additionally, we set the $\alpha$ and $\lambda$ to zero during the last 20 epochs to help with convergence, similar to what~\cite{9381661} did. We acknowledge that hyperparameter analysis is crucial. We tried different $\alpha$ values and performed a limited ablation study. The results showed that the 0.6 value results in a lower mean uncertainty value compared to the others, while making sure that the student models are not blindly copying the teacher model.

For comparison, ensemble and MC Dropout models were trained with Cross-Entropy Loss, while the self-distillation model used the loss in Equation~\ref{eq:final_dist_loss}. Consistent data folds and splits were used across methods. For self-distillation, we used $\alpha=0.6$ and $\lambda=0.04$, except for the reference layer, where we set $\alpha$ and $\lambda$ to zero for the last 20 epochs to aid convergence as suggested in~\cite{9381661}. Limited hyperparameter tuning of $\alpha$ showed that 0.6 yielded lower mean uncertainty without complete teacher copying. We acknowledge that a more in-depth ablation study is necessary.
% The details of the ablation study are reported in Appendix~\ref{apd:abl}. We plan to do a thorough hyperparameter analysis study to further study the impact of different hyperparameters on our approach.

\subsection{Results and Discussion}\label{subsec:results}
Table~\ref{tab:performance} shows the average performance, training time,  inference time (in seconds), calibration errors on test data, and number of model parameters. For the Maximum Calibration Error (MCE), we used 10 bins. MC Dropout predictions were sampled 100 times, using the mean predicted probability for decisions. The models have not been calibrated, as shown by the similar calibration errors (MCE and Brier score). All approaches perform similarly on both datasets. However, the self-distillation approach has comparable training and inference times to those of a single model, thereby reducing the computational cost of using multiple GNN models, which is particularly important for resource-limited devices. On more complex graphs, such as MIMIC-IV, execution time increases for all methods; however, self-distillation retains an inference time comparable to that of a single model. Adding a harmonization layer before each classifier in self-distillation would slightly increase complexity, but as a shallow linear layer, the approach remains more efficient than the two baselines. Table~\ref{tab:performance} also shows that the ensemble has 3x (MIMIC-IV) and 4x (Enzymes) more parameters than single models. Self-distillation and MC Dropout have comparable parameters, making them more efficient to train and use. However, MC Dropout's 100 inference samples make self-distillation more computationally efficient overall, especially for resource-constrained deployment, such as in heart-rate monitoring devices.
% Further details on efficiency and calibration error are in Tables~\ref{tab:parameters} and~\ref{tab:more_details} in Appendix~\ref{apd:more_res}.

% The average performance, time taken to train the models, and the inference time on test data (in seconds) have been included in Table~\ref{tab:performance}. We sampled the MC Dropout predictions 100 times and used the mean predicted probability for decision-making. None of the models have been calibrated. As can be seen, all approaches have similar performance on the test data for both datasets. However, the self-distillation approach takes a comparable amount of training and inference time to a single model. Therefore, using self-distillation, we can address the challenge regarding the computation cost of training and utilizing multiple GNN models, which is crucial when using the approach in resource-restricted devices. Additionally, on more complex graphs, i.e., the MIMIC-IV dataset, all approaches would take longer to execute. However, the self-distillation approach has a Test Time comparable to that of a single model. If a harmonization layer is required before each classifier of the self-distillation approach, it will slightly affect the computational complexity of the approach. Since the harmonization layer is a shallow linear layer, the approach would still be more computationally efficient than the two baselines. Table~\ref{tab:parameters,tab:more_details} in the Appendix~\ref{apd:more_res} provide more metrics and discussion on the efficiency and calibration error of the models. 

\begin{table}[b]
\centering
%\vspace{-.5cm}
    \caption{Performance, train and inference time, calibration error, and number of parameters for each method.}\label{tab:performance}
    \resizebox{1\textwidth}{!}{
        \begin{tabular}{|c|c|c|c|c|c|c|c|c|}
        \hline
             \textbf{Dataset}&\textbf{Model} & \textbf{F1-Score} & \textbf{ROC AUC} & \textbf{Train Time}&\textbf{Test Time}&\textbf{MCE}&\textbf{Brier}& \textbf{Params}\\
            \hline
             \multirow{4}{*}{MIMIC-IV}&Single Model& 0.83 $\pm$ 0.06& 0.89 $\pm$ 0.0& 293.10  $\pm$ 2.07&0.53 $\pm$ 0.07& -& -& 142,978\\
  &MC Dropout& 0.85 $\pm$  0.05& 0.89 $\pm$ 0.01
& 319.18 $\pm$ 5.09&19.19 $\pm$ 0.17& 0.11 $\pm$ 0.02& 0.11 $\pm$ 0.01& 142,978\\
  &Ensemble& 0.84 $\pm$ 0.07& 0.89 $\pm$ 0.01
& 880.71 $\pm$ 6.92&1.48 $\pm$ 0.15& 0.15 $\pm$ 0.06& 0.12 $\pm$ 0.02& 428,934\\
             &Self-Distillation& 0.88 $\pm$ 0.03& 0.90 $\pm$ 0.0& 320.04 $\pm$ 3.33&0.50 $\pm$ 0.02& 0.14 $\pm$ 0.06& 0.11 $\pm$ 0.01& 144,006\\
            \hline
            \multirow{4}{*}{Enzymes}&Single Model& 0.64 $\pm$ 0.04& 0.85 $\pm$ 0.03& 51.67 $\pm$ 2.16&0.02 $\pm$ 0.0& -& -& 73,082\\
  &MC Dropout& 0.66 $\pm$ 0.03& 0.88 $\pm$ 0.02
& 40.92 $\pm$ 1.94&1.24 $\pm$ 0.06& 0.46 $\pm$ 0.08& 0.53 $\pm$ 0.05& 73,082\\
  &Ensemble& 0.68 $\pm$ 0.03& 0.89 $\pm$ 0.02
& 204.29 $\pm$ 4.01 &0.08 $\pm$ 0.0& 0.40 $\pm$ 0.12& 0.44 $\pm$ 0.05& 292,328\\
             &Self-Distillation& 0.67 $\pm$ 0.02& 0.87 $\pm$ 0.02
& 91.78 $\pm$ 2.36 &0.02 $\pm$ 0.0& 0.42 $\pm$ 10& 0.51 $\pm$ 0.04& 88,748\\
             \hline
        \end{tabular}
        }
\end{table}

\begin{figure}[t]
\centering
    {\includegraphics[width=.9\textwidth]{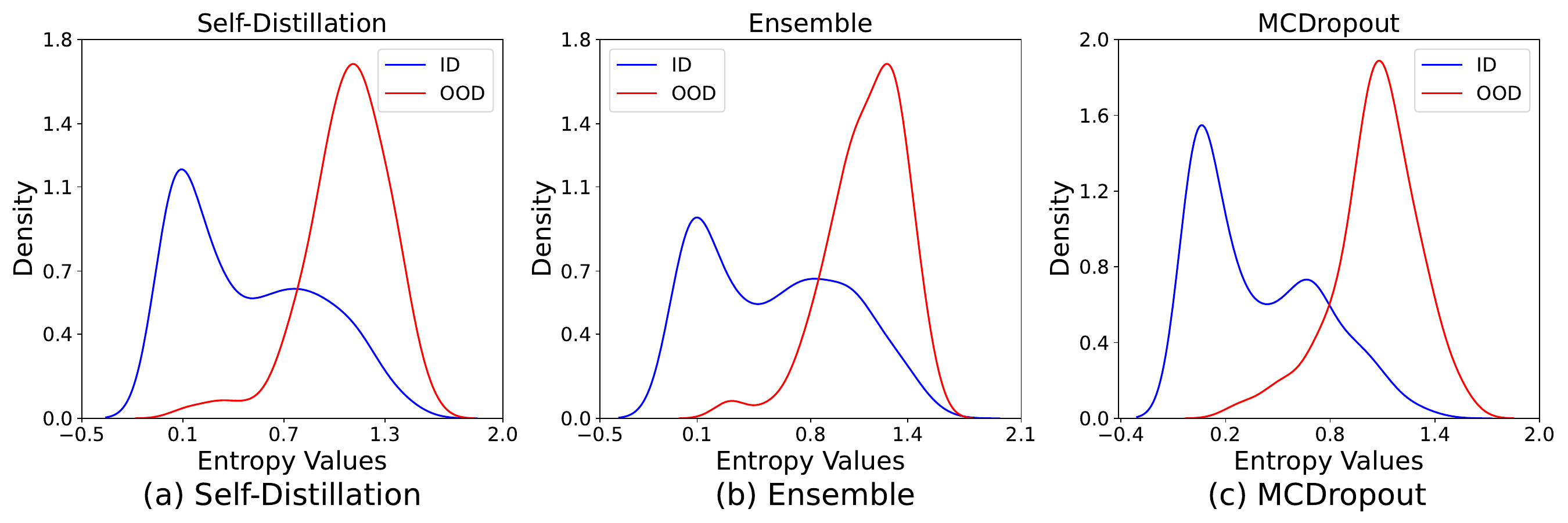}}
    \caption{KDE plots of entropy for self-distillation (a), ensembles (b), and MC Dropout (c) on Enzymes dataset. X-axis: entropy; Y-axis: probability density.
    % KDE plot of entropy values for self-distillation (a), ensemble models (b), and MC Dropout (c) on the Enzymes dataset. The X-axis shows entropy, and the Y-axis shows estimated probability density.
    }\label{fig:ensemble_vs_self_distillation_vs_mc}
\end{figure}

\vspace{.2cm}\noindent\textbf{Finding 1:} Using the proposed approach, we can efficiently train multiple GNN classifiers simultaneously, while achieving performance levels comparable to the MCDropout method and an ensemble of GNNs.
\\
\\
We evaluated trained ensemble, self-distillation, and MC Dropout models on in-distribution (ID) and out-of-distribution (OOD) Enzymes data by comparing the entropy of their predictions (Figure~\ref{fig:ensemble_vs_self_distillation_vs_mc}). Higher entropy for OOD data (the red plot in Figure~\ref{fig:ensemble_vs_self_distillation_vs_mc}) indicates successful distinction across all methods. The Enzymes dataset's six distinct classes made it suitable for visualizing this separation, unlike the less distinct classes in MIMIC-IV.

% We applied the trained ensemble, self-distillation, and MC Dropout models on OOD data, i.e., data from the holdout class, as well as the in-distribution (ID) data of the Enzymes dataset. To compare the three approaches, we evaluated their entropy of the predictive distribution. Figure~\ref{fig:ensemble_vs_self_distillation_vs_mc} shows the Kernel Density Estimation (KDE) plot for the self-distillation approach (Figure~\ref{fig:ensemble_vs_self_distillation_vs_mc}(a)), the ensemble model (Figure~\ref{fig:ensemble_vs_self_distillation_vs_mc}(b)), and the MC Dropout approach (Figure~\ref{fig:ensemble_vs_self_distillation_vs_mc}(c)) on the Enzymes dataset, where the X-axis captures the entropy values. The blue plots show the entropy of the test data, and the blue plots show the mean entropy of each OOD data. As can be seen, all approaches distinguish ID and OOD data by having higher entropy values on the OOD data. The figure shows that the approaches have higher entropy for OOD data. The Enzymes dataset was selected for this plot due to its six distinct classes. In contrast, the MIMIC-IV dataset has three classes, with 'Admitted to ICU' being a type of hospitalization, thus not fully distinct.

Self-distillation shows lower entropy on ID data than the ensemble, indicating lower uncertainty. The ensemble exhibits higher uncertainty even for ID data. This is because self-distillation trains shallower models to mimic a deeper reference model; close agreement suggests easily classified data with lower uncertainty. In contrast, independently trained ensemble members lead to higher entropy on ID data.

% In the self-distillation approach, there is a lower entropy in test data compared to the ensemble approach, which shows that the self-distillation approach has lower uncertainty for ID data. The density of higher entropy values for the test data is higher in the ensemble approach compared to the self-distillation approach. In other words, the ensemble approach is more uncertain, even regarding the ID data. This is caused by the fact that in self-distillation, the shallower models are trained to mimic the reference model. For a given data point, if a shallow model can closely mimic a reference model (which is the deepest model in our approach), then the given data point is considered to be easy to classify, i.e., the predictive uncertainty is lower. Therefore, there should be less discrepancy between the predictions of the shallow models and the reference models when the given data point is easy to classify. On the other hand, in the ensemble approach, separate models are trained separately. As a result, the outcome of the ensemble shows higher entropy for the ID data.

\vspace{.2cm}\noindent\textbf{Finding 2:} The proposed approach shows lower entropy on ID data but higher on OOD data, akin to the ensemble and MCDropout methods. The ensemble aims not to align models on class prediction, leading to slightly higher entropy on ID data than the proposed approach.
\\

Table~\ref{tab:sample_test_uncert} shows two MIMIC-IV test cases. Patient~1 has unanimous "No Admission" predictions, while Patient~2 has complete disagreement. Consequently, Table~\ref{tab:sample_test_uncert_vals} demonstrates higher uncertainty for Patient~2 using the proposed metrics, Equations~\ref{eq:uncertainty} and ~\ref{eq:uc_norm}, despite equal disagreement using Equation~\ref{eq:disagreement}, are highlighting the metric's sensitivity to prediction diversity of diverse classifiers.

The disagreement metric in Equation~\ref{eq:disagreement} fails to capture varying difficulty levels between samples, as shown in Table~\ref{tab:sample_test_uncert_vals}. While both patients have the same disagreement value, Patient~2's classifiers show a disagreement in the predicted class from the reference, indicating greater difficulty. Our proposed uncertainty metric addresses this by assigning higher disagreement values for more challenging cases. Both patients were correctly classified by the reference classifier. Furthermore, Patient~2's misclassification, unlike Patient~1's correct classification by the shallower classifiers, highlights the need for precise uncertainty metrics in healthcare to improve clinician trust and decision-making. This is especially challenging when deep learning models need to be deployed on edge devices, such as wearable heart-rate monitoring systems with limited computational resources.

\begin{table}
%\vspace{-.5cm}
\centering
\begin{minipage}[h]{0.48\textwidth}
    \centering
    \caption{Disagreement between reference and shallower classifiers for two MIMIC-IV patients.
    % Two patients from the MIMIC-IV dataset. There is a disagreement between the predicted class of the ref. and shallower classifiers in one sample.
    }
    \label{tab:sample_test_uncert}
    \resizebox{\textwidth}{!}{%
        \begin{tabular}{|c|c|c|c|c|}
        \hline
            &\multicolumn{2}{c|}{\textbf{Patient 1}}&\multicolumn{2}{c|}{\textbf{Patient 2}}\\ \hline
            \textbf{Model} & \textbf{No ADT}& \textbf{ADT}& \textbf{No ADT}& \textbf{ADT}\\ \hline
            Classifier 1 & \textit{87.89}& 12.11& \textit{57.91}& 42.09\\ 
            Classifier 2 & \textit{99.14}& 0.86& \textit{68.16}& 31.84\\ 
            Ref. Classifier& \textit{97.53}& 2.47& 47.54& \textit{52.46}\\
        \hline
        \end{tabular}
    }
\end{minipage}%
\hspace{0.03\textwidth}
\begin{minipage}[h]{0.48\textwidth}
    \centering
    \caption{MIMIC-IV sample scenario illustrating the agreement metric's lack of expressivity.
    % Sample scenario from the MIMIC-IV dataset that shows the agreement metric lacks expressivity.
    }
    \label{tab:sample_test_uncert_vals}
    \resizebox{\textwidth}{!}{%
        \begin{tabular}{|c|c|c|}
        \hline
            \textbf{Metrics}&\textbf{Patient 1}&\textbf{Patient 2}\\ \hline
            Disagreement & 0.1082& 0.1082\\ 
            UC (Linear Weight) & 0.0207& 0.0438\\ 
            UC (Nonlinear Weight) & 0.0207& 0.0498\\ 
            UC\textsubscript{norm} (Linear Weight) & 0.0099& 0.0211\\ 
            UC\textsubscript{norm} (Nonlinear Weight) & 0.0085& 0.0205\\
        \hline
        \end{tabular}
    }
\end{minipage}
\end{table}

\vspace{.2cm}\noindent\textbf{Finding 3:} The uncertainty quantification metric distinguishes scenarios with discrepancies between the teacher classifier's predicted label and shallower classifiers, making it more precise than the earlier disagreement metric.
\section{Conclusion}\label{sec:conclusion}
This paper presents an efficient and precise method to quantify predictive uncertainty in GNNs. We use self-distillation to train diverse teacher-student models in a single training process. Our approach utilizes the Jensen-Shannon divergence and a weight function to assess uncertainty, capturing disagreement between the student and teacher models. The evaluation shows performance comparable to that of the ensemble and MC Dropout methods, with reduced training and inference time. Additionally, the method identifies scenarios where student predictions diverge from a reference model.

As a future direction, we plan to explore alternative teacher selection methods (e.g., ensemble teacher distillation, transitive teacher distillation, and dense teacher distillation)~\cite{9381661} to confirm that our proposed technique for quantifying uncertainty can be applied to different self-distillation strategies. Furthermore, we aim to carry out a more thorough analysis of the weight function selection to investigate the impact of different weight functions on uncertainty quantification. Additionally, we will develop interpretable uncertainty values to improve clinical decision-making. Finally, we plan to conduct a more in-depth ablation study to investigate the impact of different values for imitation ($\alpha$) and trade-off ($\lambda$) parameters in the training of student models.
\begin{credits}
\subsubsection{\ackname}\label{sec:ack}
This study is supported by the \textit{Pediatric Mental Health Learning Health System} research project and funded by the \textit{Hamilton Health Sciences RFA Research Strategic Initiative Program} and \textit{Natural Sciences and Engineering Research Council of Canada (NSERC) Discovery} grants. This research was also undertaken thanks in part to funding from the \textit{Canada First Research Excellence Fund}. We would like to extend our deepest gratitude to Dr.~Roberto Sassi (Associate Professor and Division Head, Child \& Adolescent Psychiatry, University of British Columbia), Dr.~Paulo Pires (Psychologist and Clinical Director for the CYMHP) and Dr.~Laura Duncan (Assistant Professor, Psychiatry \& Behavioural Neurosciences, McMaster University), for their valuable insights.
\subsubsection{\discintname}
The authors have no competing interests to declare that are relevant to the content of this article.
\end{credits}

\bibliographystyle{splncs04}
\bibliography{refs}

\end{document}